%% file: main.tex
\documentclass[11pt,letterpaper]{article}

\usepackage[letterpaper,top=1in,bottom=1in,left=1in,right=1in]{geometry}
\usepackage[T1]{fontenc}
\usepackage[utf8]{inputenc}
\usepackage{amsmath}
\usepackage[charter]{mathdesign}   
\usepackage[scaled=0.95]{zi4}       
\usepackage{microtype}

\usepackage{graphicx}
\usepackage{booktabs}
\usepackage{multirow}
\usepackage{xcolor}
\usepackage{subcaption}
\usepackage[numbers,sort&compress]{natbib}
\usepackage[hidelinks,backref=page]{hyperref}  
\usepackage{csquotes}                          
\usepackage[capitalise,noabbrev]{cleveref}
\usepackage{titlesec}
\usepackage{caption}
\usepackage{enumitem}
\usepackage{makecell}
\usepackage{xurl}
\usepackage{algorithm}
\usepackage{algpseudocode}

\graphicspath{{figures/}}

\titleformat{\section}{\normalfont\large\bfseries}{\thesection}{1em}{}
\titlespacing*{\section}{0pt}{1.2ex plus .5ex minus .2ex}{0.8ex plus .2ex}
\titleformat{\subsection}{\normalfont\normalsize\bfseries}{\thesubsection}{1em}{}
\titlespacing*{\subsection}{0pt}{1ex plus .5ex minus .2ex}{0.6ex plus .2ex}
\titleformat{\paragraph}[runin]{\normalfont\normalsize\bfseries}{}{0pt}{}
\titlespacing*{\paragraph}{0pt}{0.6ex plus .3ex}{1em}

\setlist{nosep,leftmargin=*}



\definecolor{topicnews}{HTML}{F8DDA6}
\definecolor{topicservice}{HTML}{A9DCCB}
\definecolor{topicenterprise}{HTML}{A6CBE3}
\newcommand{\topicnews}[1]{{\setlength{\fboxsep}{0.6pt}\colorbox{topicnews}{\strut #1}}}
\newcommand{\topicservice}[1]{{\setlength{\fboxsep}{0.6pt}\colorbox{topicservice}{\strut #1}}}
\newcommand{\topicenterprise}[1]{{\setlength{\fboxsep}{0.6pt}\colorbox{topicenterprise}{\strut #1}}}

\newcommand{\papertitle}{Train on prefill, monitor at decode:\\ Using sparse autoencoders for topic monitoring}

\hypersetup{pdftitle={Train on prefill, monitor at decode: Using sparse autoencoders for topic monitoring},
            pdfauthor={Rishikesh Mallagundla, Pranav Patel}}

\begin{document}

\begin{center}
  \rule{\textwidth}{4pt}\par\vspace{0.12in}
  {\LARGE\bfseries \papertitle\par}
  \vspace{0.1in}\rule{\textwidth}{1pt}\par
  \vspace{0.25in}
  \begin{tabular}[t]{c}
    \textbf{Rishikesh Mallagundla}\\ Mercity\\ \texttt{rishikesh@mercity.io}
  \end{tabular}\hspace{0.6in}%
  \begin{tabular}[t]{c}
    \textbf{Pranav Patel}\\ Mercity\\ \texttt{pranav@mercity.io}
  \end{tabular}
  \par\vspace{0.14in}
  \href{https://www.mercity.ai}{\includegraphics[height=0.4in]{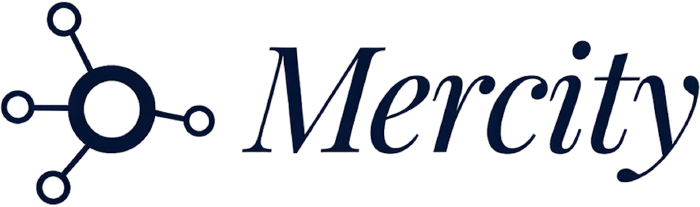}}
  \par\vspace{0.25in}

  {\large\bfseries Abstract\par}
  \vspace{1.5ex}
  \begin{minipage}{0.88\textwidth}
    \input{sections/00_abstract}
  \end{minipage}
\end{center}
\vspace{0.2in}

\input{sections/01_introduction}
\input{sections/02_related_work}
\input{sections/02b_preliminaries}
\input{sections/03_methodology}
\input{sections/05_results}
\input{sections/06_discussion}

{\small
\bibliographystyle{plainnat}
\bibliography{references}
}

\appendix
\section*{\Large Appendix}
\input{sections/appendix}

\end{document}

%% file: sections/00_abstract.tex
Topic monitoring is the task of labelling the topic of each token in a model's response during generation. Activation-based monitors that run at generation time have so far been trained to detect harmful content and are limited to two or three classes. We introduce TopicScope, a framework for training causal topic classifiers on SAE features that assign a topic to every token, so that the label can change partway through a response. We train three detector architectures on synthetic multi-topic conversations covering seven topics. Across Gemma-3 1B and 4B, monitors trained on features extracted from prefill hidden states were able to generalise to decoding. We also found that the classifiers are robust to perturbations, losing at most 5 macro-F1 points when off-topic asides are inserted, topics are removed or sections are reordered. Ablations conducted to understand the contribution of the SAE revealed that classifiers reaching baseline performance without it were several times larger. Our findings show that compact topic monitors can be trained inexpensively in prefill and deployed during generation. More broadly, our work shows where SAEs are useful beyond safety classification. Our code, dataset and checkpoints are available on \href{https://github.com/Mercity-AI/SAE-Guardrails}{\textbf{GitHub}} and \href{https://huggingface.co/collections/Mercity/online-guardrails-using-sparse-autoencoders-6ac602940468c62fefb2b9e8}{\textbf{Hugging Face}}.

%% file: sections/01_introduction.tex
\section{Introduction}
\label{sec:intro}

Language models are frequently deployed as single-purpose assistants, for tasks such as customer support, billing or clinical triage. Most deployed assistants serve a fixed purpose and any kind of unintentional drift during response generation should be avoided to save costs and align the model to its mission. For example, a billing assistant for a telecom company should talk about invoices, payments and refunds, and avoid topics such as tax or legal advice. Even though this topic list can be ingrained in the model through a carefully designed system prompt or through fine-tuning \citep{sreedhar2024canttalkaboutthis}, the model can still drift off topic over the course of a multi-turn conversation \citep{rebedea2023nemo,sreedhar2024canttalkaboutthis}.

Topic monitoring is the task of labelling the topic of each token in a model's response during generation, which downstream can enable catching the point where a response leaves its scope while it is still being written.

Monitoring a response while it is being written is already common practice for safety. Guard models such as Llama Guard \citep{inan2023llama} and ShieldGemma \citep{zeng2024shieldgemma} classify a prompt or a finished response, and streaming guardrails extend them to label the response token by token as it is generated \citep{zhao2025qwen3guard,li2025judgment,yu2026sentguard}. A parallel line of work reads the generating model's own activations instead of its text. Linear probes on these activations screen for jailbreaks \citep{cunningham2026constitutional} and detect deception \citep{goldowskydill2025detecting} and high-stakes interactions \citep{mckenzie2025detecting}, and probes on sparse autoencoder features have flagged personal data token by token in production \citep{nguyen2025rakuten}. Across both lines of work, the label is almost always a safety judgement between benign and harmful, with two or three classes.

Topical guardrails, in turn, classify the prompt or the finished response \citep{rebedea2023nemo,chua2024flexible}, so a change of topic is caught only after it has been written, and a single label for the response may not show where the change happened. While prior work monitors safety during generation and topic after it, no existing monitor labels the topic of a response token by token as it streams. Extending streaming safety monitors from two or three classes to many topics, read from the generating model's own features, would close this gap.

We introduce TopicScope, a framework for training topic monitors using sparse autoencoders (SAEs). TopicScope relies on the fact that, in a causal language model, the hidden state at a token is computed the same way whether that token was read in prefill or generated in decode. We therefore pass conversations annotated with topics at the token level through the backbone in prefill, train small causal detectors on the SAE features of the resulting hidden states, and then run the same detectors unchanged on the hidden states the backbone produces while it generates.

We use TopicScope to answer four research questions, each building on the one before it: \textbf{(RQ1)} Do SAE features carry a per-token topic signal? \textbf{(RQ2)} Do detectors trained in prefill transfer to decode? \textbf{(RQ3)} Do the detectors stay accurate when the response is perturbed? \textbf{(RQ4)} Is the SAE necessary?

Our core contributions are:

\begin{enumerate}
  \item We train causal topic monitors on SAE features for Gemma-3 1B and 4B \citep{team2025gemma}, using synthetic multi-topic conversations over seven topics and three detector architectures: a GRU, a ConvNeXt-style temporal convolutional network and a Transformer. We release the synthetic conversation dataset along with the code that generates it and the trained detector checkpoints on \href{https://github.com/Mercity-AI/SAE-Guardrails}{GitHub} and \href{https://huggingface.co/collections/Mercity/online-guardrails-using-sparse-autoencoders-6ac602940468c62fefb2b9e8}{Hugging Face}.
  \item We show that these monitors, trained only on teacher-written text in prefill, carry over to decode and follow the topics of the responses the backbone generates itself. They maintain their scores even when we insert off-topic text, remove a topic or reorder the sections of a response.
  \item We train the same detectors on raw hidden states to test whether the SAE is needed, and find that its benefit is a monitor whose size does not grow with the backbone. The raw-state detectors reach the same performance only once they are several times larger.
\end{enumerate}

The rest of the paper is organised as follows: \cref{sec:related} reviews prior work on topic monitoring and sparse autoencoders, \cref{sec:prelim} introduces the language models, sparse autoencoders and causal detectors we build on, \cref{sec:method} outlines TopicScope's architecture and the dataset generation pipeline, \cref{sec:results} answers the research questions and discusses the findings, and \cref{sec:discussion} concludes with directions for future work.

%% file: sections/02_related_work.tex
\section{Background and Related Work}
\label{sec:related}

\paragraph{Topic modelling and segmentation.} Topic models summarise a corpus as a set of themes. The classic approach is a bag-of-words generative model such as LDA \citep{blei2003lda}, later followed by neural variants \citep{srivastava2017autoencoding}, clustering of document embeddings \citep{grootendorst2022bertopic} and prompted LLMs \citep{pham2023topicgpt}. Topic segmentation splits a single document into topically coherent parts, either from lexical cohesion \citep{hearst1997texttiling} or as a supervised sequence labelling task \citep{koshorek2018text}. In dialogue, topic-shift detection asks whether the topic changes at a given utterance \citep{lin2023multi}. All of these methods work on text that has already been written. Topic models discover their themes from a whole collection, and segmentation places boundaries after the fact. TopicScope instead assigns one of a fixed set of topics to every token while the response is still being generated.

\paragraph{SAEs as topic models.} Three recent papers link SAEs to topic modelling, and all three stay unsupervised and corpus-level like the models above. \citet{girrbach2025sparse} derive the SAE objective as a maximum a posteriori estimator under a continuous topic model inspired by LDA. Mechanistic Topic Models \citep{zheng2025directions} and MonoTM \citep{joh2026top} describe topics with SAE features in place of word lists, which lets a topic capture ideas that a list of top words cannot.

\paragraph{Topical guardrails.} Most scope control in deployed systems works on the text. NeMo Guardrails \citep{rebedea2023nemo} lets developers write programmable rails over a dialogue, including rails that keep the conversation away from given topics. Such rails depend on curated examples, and \citet{chua2024flexible} remove that dependence by generating synthetic prompts from a description of the intended scope, then training classifiers that judge whether a user prompt is relevant to the system prompt. Fence \citep{lakara2026fence} trains similar specialised small models for other application-specific checks, topic drift among them. Instead of screening the conversation, CantTalkAboutThis \citep{sreedhar2024canttalkaboutthis} fine-tunes the model itself to stay on topic. More recent methods move the check from the text to the model's activations. \citet{pan2026conversational} fit a one-class SVM on hidden states to flag conversation turns that fall out of context, and kNNGuard \citep{abdelfattah2026knnguard} classifies prompts by their nearest neighbours in the activation space of an off-the-shelf model. Whether they read text or activations, all of these methods make one decision per prompt, turn or response, and none labels a response while it is being generated.

\paragraph{Safety moderation and streaming guardrails.} Moderation models such as Llama Guard \citep{inan2023llama}, ShieldGemma \citep{zeng2024shieldgemma} and WildGuard \citep{han2024wildguard} classify a prompt or a complete response against a harm taxonomy. Streaming guardrails move that check into generation. Stream Qwen3Guard adds a token-level head that labels output as it arrives \citep{zhao2025qwen3guard}. SCM trains a token-level monitor on heuristic per-token harm labels and stops a response once enough tokens are flagged \citep{li2025judgment}. SentGuard moderates at sentence boundaries to avoid unstable token-level decisions \citep{yu2026sentguard}, while FreoStream and StreamGuard try to anticipate harm before it is fully written \citep{wang2026freostream,kavumba2026predict}. In timing, these systems are the closest to ours. They are separate models, often several billion parameters, that read the generated text, and their labels are safety classes with two or three values. TopicScope, by contrast, reads the generating model's own activations and separates many topics.

\paragraph{Activation probes as monitors.} Linear probes were introduced to study what a layer encodes \citep{alain2016understanding,belinkov2021probing}, and representation engineering extended them to reading and steering high-level concepts \citep{zou2023representation}. They are now used as runtime monitors. Constitutional Classifiers++ uses a probe as a cheap first stage before a larger classifier \citep{cunningham2026constitutional}. Probes also detect strategic deception \citep{goldowskydill2025detecting}, high-stakes interactions \citep{mckenzie2025detecting} and reward hacking \citep{bergen2026rewardhacking}. In method, this line of work is the closest to ours. Each probe, though, targets a single binary property, and most score a whole prompt or response. \citet{mckenzie2025detecting} also find that probes trained on synthetic data generalise to real interactions, which is the assumption behind our teacher-written training text.

\paragraph{SAE features as classifier inputs.} Whether SAE features make better classifier inputs than raw activations is contested. \citet{kantamneni2025sparse} test SAE probes under data scarcity, class imbalance, label noise and covariate shift and find no consistent gain over baselines. \citet{smith2025negative} report SAE probes trailing linear probes on out-of-distribution harmful-intent detection, and AxBench reaches a similar conclusion for steering and concept detection \citep{wu2025axbench}. Other results are more positive. \citet{gallifant2025sparse} find SAE features outperforming hidden-state baselines on safety classification and transferring from Gemma 2 2B to 9B, and SAE probes trained on synthetic data detected personal data in production at Rakuten, transferring to real traffic better than other probing methods \citep{nguyen2025rakuten}. Practitioner guidance recommends against an SAE when the target concept is known in advance \citep{goodfire2026guide}. Topics are such a concept, and our ablation without the SAE tests this guidance on a multi-class, per-token task.

%% file: sections/02b_preliminaries.tex
\section{Preliminaries}
\label{sec:prelim}

\paragraph{Decoder-only language models} such as GPT \citep{radford2018improving,radford2019language} and Gemma 3 \citep{team2025gemma} are trained to predict each token from the ones before it. Given a sequence of tokens $x_1, \dots, x_T$, such a model factorises the probability of the sequence as
\begin{equation}
  p_\theta(x_1, \dots, x_T) = \prod_{t=1}^{T} p_\theta(x_t \mid x_{<t}),
\end{equation}
and is trained to maximise the log-likelihood of a large text corpus under this factorisation,
\begin{equation}
  \mathcal{L}(\theta) = \sum_{t=1}^{T} \log p_\theta(x_t \mid x_{<t}).
\end{equation}
To make each of these predictions, the model computes a hidden state $h_t^{(\ell)}$ at every position $t$ in each layer $\ell$ of a Transformer stack \citep{vaswani2017attention}. These hidden states accumulate in the residual stream, which carries everything later layers use to predict the next token. Since the objective conditions each prediction on $x_{<t}$ alone, attention is masked to earlier positions, and the hidden state at position $t$ depends only on $x_1, \dots, x_t$.

This causal dependence holds in both ways a model processes text. During prefill, the model reads a given prompt in one forward pass, and during decode it generates one token at a time, appending each new token to the context and reusing the cached keys and values of the earlier ones.

\paragraph{Sparse autoencoders (SAEs)} are dictionary-learning models that rewrite a language model's activations as a sparse combination of a large number of features \citep{bricken2023monosemanticity,cunningham2023sparse}. Given a residual-stream activation $h \in \mathbb{R}^d$, an SAE encodes it into a feature vector
\begin{equation}
  f(h) = \sigma(W_{\text{enc}} h + b_{\text{enc}}) \in \mathbb{R}^m, \qquad m \gg d,
\end{equation}
and reconstructs the activation from those features as
\begin{equation}
  \hat{h} = W_{\text{dec}} f(h) + b_{\text{dec}}.
\end{equation}
Training penalises the number of active features, so only a small fraction of the $m$ features fire on a given token, and many of them respond to a single recognisable concept. The Gemma Scope 2 SAEs \citep{gemmascope2} are trained on the residual stream of every layer of Gemma 3. They use the JumpReLU activation \citep{rajamanoharan2024jumping},
\begin{equation}
  \sigma(z) = z \cdot \mathbb{1}[z > \theta],
\end{equation}
which passes a pre-activation through unchanged when it exceeds a learned per-feature threshold $\theta$ and sets it to zero otherwise.

\paragraph{Causal detectors.} A causal detector is a sequence model that labels each position of an input sequence $z_1, \dots, z_T$ using only that position and the ones before it,
\begin{equation}
  \hat{y}_t = g(z_1, \dots, z_t).
  \label{eq:causal}
\end{equation}
Such a detector can emit the label for a token as soon as the token is produced, which is what a monitor running alongside decode requires, since the tokens after it do not exist yet. Recurrent, convolutional and attention-based models meet this constraint in different ways, and each makes a different assumption about how much of the past a label needs.

\paragraph{A gated recurrent unit (GRU)} \citep{cho2014learning} keeps a hidden state $s_t$ that it updates once per token from the previous state and the current input,
\begin{equation}
  s_t = \mathrm{GRU}(s_{t-1}, z_t),
\end{equation}
with gates that decide how much of the old state to keep and how much to overwrite. Because $s_t$ depends only on $s_{t-1}$ and $z_t$, the GRU is causal by construction and needs no mask or padding. The same recurrence means each new token costs one state update, no matter how long the sequence has grown. The whole history, however, has to fit in a fixed-size state, so a topic set many tokens earlier survives only as long as the gates keep it.

\paragraph{A temporal convolutional network (TCN)} \citep{bai2018empirical} labels each token from a window of recent inputs instead of a running state. A one-dimensional convolution becomes causal when its input is padded on the left only, so the output at $t$ covers positions $t-k+1, \dots, t$ for a kernel of size $k$. Stacking such layers with dilations that double at each layer widens this window exponentially with depth. ConvNeXt blocks \citep{liu2022convnet} can serve as these layers. Each block pairs a depthwise convolution with a per-token MLP and a residual connection, and since the depthwise convolution is the only part that mixes positions, padding it on the left is enough to make the whole block causal.

\paragraph{A Transformer encoder} \citep{vaswani2017attention} lets the representation at position $t$ attend directly to every earlier position. Applying the same lower-triangular mask that a decoder-only language model uses blocks attention to later positions, and this mask alone makes the encoder causal. Where the GRU compresses the past into a state and the TCN restricts it to a fixed window, the Transformer keeps the full history available at every token, at a cost that grows with the length of the sequence.

%% file: sections/03_methodology.tex
\section{Methodology}
\label{sec:method}

\subsection{Overview of TopicScope}
\label{sec:overview}

\begin{figure}[htbp]
  \centering
  \includegraphics[width=\textwidth]{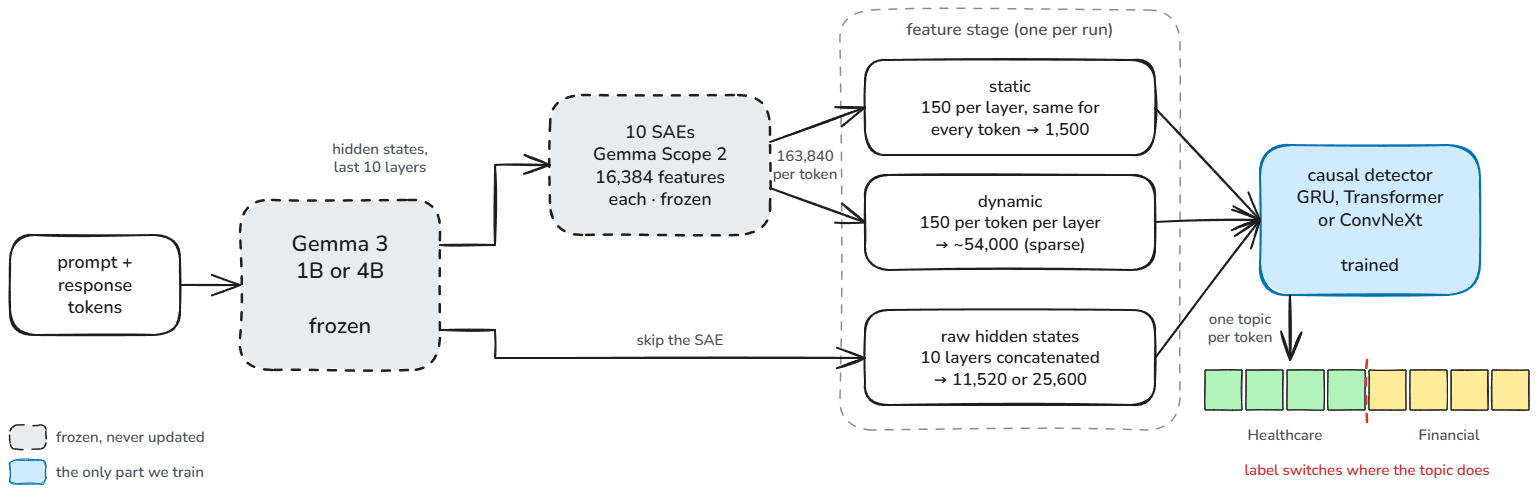}
  \caption{The TopicScope pipeline. Gemma 3 and its SAEs are frozen, a feature stage reduces the SAE output at ten layers to the detector's input, and only the causal detector is trained. The detector emits one topic per token, so its label can switch partway through a response.}
  \label{fig:pipeline}
\end{figure}

TopicScope trains a topic monitor in prefill and runs it in decode (\cref{fig:pipeline}). The pipeline has four steps:

\begin{enumerate}
  \item \textbf{Data.} A teacher model writes a corpus $\boldsymbol{\mathcal{D}}$ of $|\mathcal{D}| = 10{,}000$ conversations, each divided into topic-labelled spans in both the prompt and the response.
  \item \textbf{Prefill.} Each conversation passes through a backbone language model $\boldsymbol{\mathcal{M}}$ in a single forward pass, and we read the residual stream at its last $\boldsymbol{N}$ layers.
  \item \textbf{Features.} At each of these layers, a pretrained SAE $\boldsymbol{\mathcal{S}}$ encodes the residual stream, and a selection step reduces the encoded features to a fixed-width input $z_t$ for every token.
  \item \textbf{Detection.} A small causal detector learns to predict the topic of every token from $z_t$. At deployment, $\mathcal{M}$ writes its own response, and the same detector, unchanged, labels each token as it is generated.
\end{enumerate}

Only the detector is trained; the LLM and the SAEs stay frozen.

\subsection{Task formulation}
\label{sec:task}

Given a conversation tokenised as $x_1, \dots, x_T$, made of $\boldsymbol{T_p}$ prompt tokens followed by $\boldsymbol{T_r}$ response tokens under the backbone's chat template, topic monitoring assigns each token a label $y_t \in \mathcal{Y}$ from a set of topics $\mathcal{Y}$. We use $|\mathcal{Y}| = 7$ business topics (\cref{sec:data}). A single response may cover several of these in sequence, so the label can change partway through, and the monitor has to place that change at the token where it happens. As in \cref{sec:prelim}, the monitor is causal and predicts
\begin{equation}
  \hat{y}_t = g(z_1, \dots, z_t),
\end{equation}
where $z_t$ is the feature vector read from the backbone at token $t$. Tokens in both $T_p$ and $T_r$ carry labels when they fall inside a topic span.

The monitor is trained in prefill, where a finished conversation passes through the backbone in one forward pass and every token has a label, and deployed in decode, where the backbone writes one token at a time and no labels exist.

\subsection{Dataset}
\label{sec:data}

The corpus is generated with Simula \citep{simula2026}, our synthetic data generation library, using GPT-5.6 Luna as both generator and critic. For each sample, Simula draws a mix from a custom taxonomy built on the seven topics (Enterprise documents, General news \& content, Customer service, Legal, Financial, HR \& people operations, and Healthcare), writes a short brief describing the record to create, and generates the record from that brief. A critic then checks every record against its brief and the labelling rules, and rejected records are refined or discarded.

\begin{figure}[htbp]
  \centering
  \includegraphics{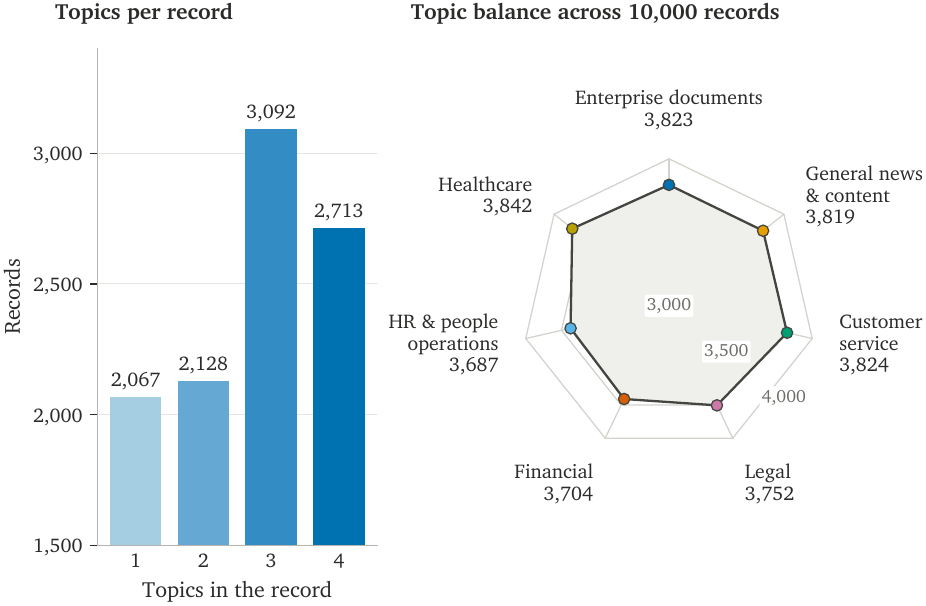}
  \caption{Composition of the corpus: records by number of topics (left) and records containing each topic (right). The bar axis starts at 1,500 records and the spider chart's centre is 3,000 records, not zero.}
  \label{fig:dataset}
\end{figure}

Each record is a single prompt--response pair, in which a short user prompt asks about one or more topics and an assistant response answers them in order. The topics and their order are fixed by the sampled mix before the record is written, and the generator marks each topic's span in both the prompt and the response, which gives every token its label (\cref{fig:sample}). Across the corpus, most pairs cover more than one topic, the seven topics are balanced, and topic sequences are sampled so that every ordered pair of topics can occur (\cref{tab:dataset,fig:dataset}).

\begin{figure}[htbp]
  \centering
  \fbox{\begin{minipage}{0.97\textwidth}
    \raggedright
    \vspace{0.4ex}
    \textbf{Prompt}\par\vspace{0.6ex}
    \topicnews{What are the main developments in recent climate and technology news?}\hspace{0.3em}\topicservice{Also help me resolve a}\\
    \topicservice{duplicate charge on my account.}\hspace{0.3em}\topicenterprise{Finally, turn the relevant operational lessons into a concise}\\
    \topicenterprise{internal memo.}

    \vspace{1.2ex}
    \textbf{Response} (excerpt)\par\vspace{0.6ex}
    \topicnews{Recent climate coverage has focused on a combination of record-setting heat, rapid investment in}\\
    \topicnews{clean energy, and the continuing difficulty of reducing emissions from heavy industry.} [\ldots] \topicnews{Across}\\
    \topicnews{both areas, the news points to fast-moving change accompanied by practical questions about}\\
    \topicnews{infrastructure, accountability, and public policy.}\hspace{0.3em}\topicservice{A duplicate charge is usually best handled by first}\\
    \topicservice{comparing the transaction details rather than assuming that two entries represent the same}\\
    \topicservice{payment.} [\ldots] \topicservice{Review the account afterward to make sure the order status and any refund are}\\
    \topicservice{both recorded accurately.}\hspace{0.3em}\topicenterprise{The operational lesson is to make payment incidents visible early}\\
    \topicenterprise{without treating every temporary authorization as a confirmed error.} [\ldots]

    \vspace{1.2ex}
    {\centering\small \topicnews{General news \& content} \quad \topicservice{Customer service} \quad \topicenterprise{Enterprise documents}\par}
    \vspace{0.4ex}
  \end{minipage}}
  \caption{A test record that covers three topics. Each highlight marks a topic's labelled span in the prompt and the response, and the response is excerpted around its two topic changes.}
  \label{fig:sample}
\end{figure}

\begin{table}[htbp]
  \centering
  \begin{tabular*}{\textwidth}{@{\extracolsep{\fill}}lrlr@{}}
    \toprule
    \multicolumn{2}{c}{Size} & \multicolumn{2}{c}{Topics} \\
    \cmidrule(r){1-2}\cmidrule(l){3-4}
    Records & 10,000 & Mean topics per record & 2.65 \\
    Train / validation / test & 7,000 / 1,500 / 1,500 & Records with 1 topic & 2,067 \\
    Total tokens & 8.85M & Records with 2 topics & 2,128 \\
    Labelled response tokens & 8.2M & Records with 3 topics & 3,092 \\
    Labelled prompt tokens & 515K & Records with 4 topics & 2,713 \\
    Mean prompt length & 44 words & Records containing each topic & 3,687 to 3,842 \\
    Mean response length & 675 words & & \\
    \midrule
    \multicolumn{2}{l}{Pivot style (share of multi-topic records)} & \multicolumn{2}{r}{Explicit 40\%, light 35\%, unmarked 25\%} \\
    \bottomrule
  \end{tabular*}
  \caption{The synthetic topic-shift corpus.}
  \label{tab:dataset}
\end{table}

Generation also varies how each topic change is written, so that a monitor cannot find a boundary from surface cues alone. Every multi-topic pair uses one pivot style for all its boundaries: an explicit connective such as \enquote{On a separate note}, a light or neutral connective, or an unmarked pivot that starts the new topic directly or reaches it through a person, company or object the two passages share.

\subsection{Feature extraction}
\label{sec:features}

We use Gemma 3 1B and 4B as backbones \citep{team2025gemma}, and their corresponding Gemma Scope 2 SAEs at the last $N = 10$ layers (16--25 for 1B and 24--33 for 4B) to extract features from the residual stream.

\begin{figure}[htbp]
  \centering
  \makebox[\textwidth][c]{\includegraphics[width=1.1\textwidth]{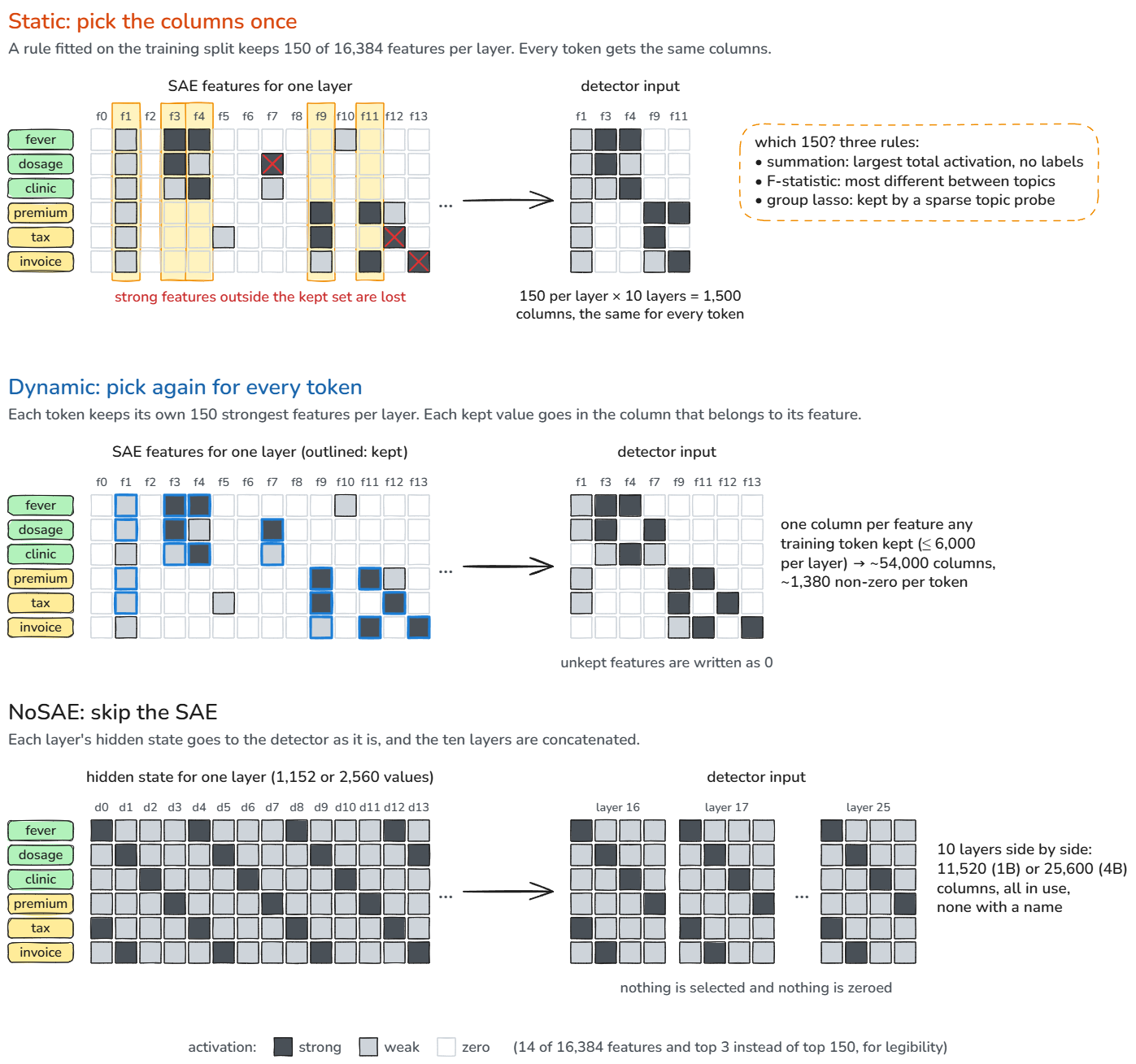}}
  \caption{Static selection keeps the same columns for every token, so a token whose strongest features fall outside them loses those features. Dynamic selection keeps each token's own strongest features and writes each into the column that belongs to it. NoSAE skips the SAE and passes the hidden states directly.}
  \label{fig:features}
\end{figure}

For efficiency, we keep only $K = 150$ features per layer and concatenate them across the $N$ layers into the detector's input $z_t$ (\cref{fig:features}), using one of the following methods to select them:

\begin{itemize}
  \item \textbf{Static selection} fixes one set of $K$ features per layer, fitted on the training split and shared by every token, using one of three rules for ranking features:
  \begin{itemize}
    \item \textbf{Summation} keeps the $K$ features with the largest summed activation over the training tokens.
    \item \textbf{F-statistic} keeps the $K$ features whose mean activation differs most between topics relative to their spread within a topic.
    \item \textbf{Group lasso} fits a linear probe from all $m$ features to the seven topics with a sparse group-lasso penalty \citep{yuan2006grouplasso}.
  \end{itemize}
  \item \textbf{Dynamic selection} keeps the top-$K$ features of each token at every layer, each written into a fixed column for that feature, so the set of active columns changes from token to token.
  \item \textbf{NoSAE}, our baseline, bypasses the SAE altogether and sends the hidden states of the $N$ layers to the detector directly.
\end{itemize}

\subsection{Detectors and training}
\label{sec:detectors}

We train the three causal detectors from \cref{sec:prelim}, (1) a \textbf{GRU}, (2) a \textbf{ConvNeXt}-style temporal convolutional network and (3) a causal \textbf{Transformer} encoder, on every feature input with token-level cross-entropy over the labelled tokens, using AdamW \citep{loshchilov2017decoupled} for 15 epochs, and keep the checkpoint with the best validation macro-F1. Architectures, parameter counts and training hyperparameters are listed in \cref{app:detectors}.

%% file: sections/05_results.tex
\section{Results}
\label{sec:results}

Every score below comes from the 1,500 held-out test records and is a percentage. Where tokens carry labels, we report \textbf{macro-F1}, the token-level F1 averaged equally over the seven topics, and \textbf{overlap}, the intersection-over-union of predicted and true tokens per topic averaged over topics. Overlap is the stricter of the two, because it counts each correct token once where F1 counts it twice.

\subsection{RQ1: Do SAE features carry a per-token topic signal?}
\label{sec:rq1}

We train the three detectors on every feature input for both backbones and score them in prefill (\cref{tab:rq1}), and report per-topic scores, the effect of supervising prompt tokens, and layer knockout and attribution in \cref{app:prefill,app:knockout}.

\begin{table}[htbp]
  \centering
  \begin{tabular*}{\textwidth}{@{\extracolsep{\fill}}llcccccc@{}}
    \toprule
     & & \multicolumn{3}{c}{Macro-F1 (\%)} & \multicolumn{3}{c}{Overlap (\%)} \\
    \cmidrule(lr){3-5}\cmidrule(l){6-8}
     & Input & GRU & ConvNeXt & Transformer & GRU & ConvNeXt & Transformer \\
    \midrule
    \multirow{5}{*}{Gemma 3 1B} & Summation & 94.5 & 96.2 & 94.3 & 89.6 & 92.6 & 89.3 \\
     & F-statistic   & 96.4 & 96.2 & 95.7 & 93.0 & 92.8 & 91.8 \\
     & Group lasso  & \textbf{96.8} & 96.3 & 95.5 & \textbf{93.8} & 92.9 & 91.5 \\
     & Dynamic & 96.1 & \textbf{97.3} & 94.0 & 92.5 & \textbf{94.8} & 88.7 \\
     & NoSAE  & 96.5 & 96.7 & \textbf{95.9} & 93.3 & 93.6 & \textbf{92.2} \\
    \midrule
    \multirow{5}{*}{Gemma 3 4B} & Summation & 87.4 & 92.4 & 91.8 & 77.7 & 86.0 & 85.0 \\
     & F-statistic   & \textbf{96.4} & 95.5 & \textbf{95.0} & \textbf{93.1} & 91.5 & \textbf{90.5} \\
     & Group lasso  & 96.3 & 95.5 & 94.9 & 92.9 & 91.4 & 90.3 \\
     & Dynamic & 92.4 & 97.1 & 94.0 & 86.0 & 94.5 & 88.7 \\
     & NoSAE  & 96.3 & \textbf{97.5} & 94.9 & 92.9 & \textbf{95.1} & 90.4 \\
    \bottomrule
  \end{tabular*}
  \caption{Test macro-F1 and overlap for every input. The NoSAE rows are discussed in \cref{sec:rq4}.}
  \label{tab:rq1}
\end{table}

\textbf{On Gemma 3 1B, even the unsupervised summation rule reaches 94.3--96.2 macro-F1 from 1,500 of the 163,840 available features, and selecting features per token (dynamic selection) adds at most 1.1 points.} This parity does not hold on Gemma 3 4B, where every summation detector scores below its Gemma 3 1B counterpart and GRU drops to 87.4.

Holding the budget at $K = 150$ features per layer and changing only the selection rule removes this drop. Both label-based rules bring every Gemma 3 4B detector to within a point of its Gemma 3 1B score and raise the macro-F1 of the Gemma 3 4B GRU from 87.4 to 96.4. The two supervised rules are within 0.4 points of each other in every row, so the F-statistic, a single deterministic pass with no hyperparameters, is the cheaper choice.

Among the detectors, ConvNeXt is the strongest in six of the ten rows, and GRU depends most on the input, spanning 87.4 to 96.8.

\subsection{RQ2: Do detectors trained in prefill transfer to decode?}
\label{sec:rq2}

The detectors from RQ1 then label, without retraining, responses that the backbone generates itself for 1,000 test prompts with the topic tags removed. These responses have no per-token labels, so each is scored against the topics its prompt requested:
\begin{itemize}
  \item \textbf{Purity:} the share of response tokens labelled with a requested topic.
  \item \textbf{Coverage:} the share of requested topics that each receive at least a tenth of the tokens.
  \item \textbf{Full coverage:} the share of responses in which every requested topic clears that bar.
\end{itemize}
These measures test consistency with the request. A low score can come from the detector mislabelling the response or from a response that strayed from the request. To separate the two, the same detectors also read the teacher's responses to the same prompts. Those responses cover every requested topic, so their scores serve as the reference.

\begin{table}[htbp]
  \centering
  \begin{tabular*}{\textwidth}{@{\extracolsep{\fill}}llcccccc@{}}
    \toprule
     & & \multicolumn{3}{c}{Purity (\%)} & \multicolumn{3}{c}{Full coverage (\%)} \\
    \cmidrule(lr){3-5}\cmidrule(l){6-8}
     & Input & GRU & ConvNeXt & Transformer & GRU & ConvNeXt & Transformer \\
    \midrule
    \multirow{5}{*}{Gemma 3 1B} & Summation & 91.1 & 89.9 & 94.5 & 68.1 & 64.3 & 66.9 \\
      & F-statistic & 93.2 & 92.2 & 96.5 & 62.7 & 73.0 & 69.3 \\
      & Group lasso & \textbf{96.6} & 93.8 & \textbf{97.7} & 69.7 & 75.6 & 67.8 \\
      & Dynamic & 94.5 & 94.4 & 94.1 & 70.4 & 77.1 & \textbf{73.1} \\
      & NoSAE & 95.8 & \textbf{95.1} & 94.7 & \textbf{77.1} & \textbf{78.5} & 63.7 \\
    \midrule
    \multirow{5}{*}{Gemma 3 4B} & Summation & 69.4 & 75.8 & 90.2 & 49.5 & 33.7 & 69.1 \\
      & F-statistic & 95.3 & 90.4 & \textbf{96.3} & 82.2 & 81.0 & 76.7 \\
      & Group lasso & \textbf{96.7} & 92.5 & 94.0 & 84.1 & 82.1 & 69.9 \\
      & Dynamic & 94.2 & 94.5 & 94.2 & 77.2 & 80.3 & 73.0 \\
      & NoSAE & 95.3 & \textbf{95.1} & 88.3 & \textbf{85.8} & \textbf{82.6} & \textbf{80.8} \\
    \bottomrule
  \end{tabular*}
  \caption{Decode scores on the backbone's own responses.}
  \label{tab:decode}
\end{table}

\textbf{The detectors transfer.} Gemma's responses differ from the teacher's, averaging about 1,240 tokens against 820, yet with dynamic or label-selected features 90--98\% of its tokens receive a requested topic, at both sizes and for all three detectors (\cref{tab:decode}). The exception is summation on Gemma 3 4B, the weakest input in prefill, where the gap widens. The Gemma 3 4B GRU trailed dynamic selection by 5 macro-F1 points on teacher text but trails it by 25 purity points on Gemma's. Label-based selection corrects this, raising GRU's purity from 69 to 95 and ConvNeXt's full coverage from 34 to 81.

Full coverage is the weakest decode score (\cref{tab:teacher}). On the teacher's text the same detectors find every topic the text contains, reaching 95--99 full coverage. On Gemma's text it falls by about 30 points on Gemma 3 1B and further on Gemma 3 4B, and it drops as prompts request more topics, from near 100 on single-topic prompts to 34--62 on four-topic prompts. Raising the generation limit from 1,500 to 8,000 tokens seems to not change the scores, as Gemma's responses end well before either limit. Most of the coverage gap therefore comes from Gemma, which answers fewer of the topics in a crowded prompt.

\begin{table}[htbp]
  \centering
  \begin{tabular*}{\textwidth}{@{\extracolsep{\fill}}llcccc@{}}
    \toprule
     & & \multicolumn{2}{c}{Purity (\%)} & \multicolumn{2}{c}{Full coverage (\%)} \\
    \cmidrule(lr){3-4}\cmidrule(l){5-6}
     & & Teacher & Gemma & Teacher & Gemma \\
    \midrule
    \multirow{3}{*}{Gemma 3 1B} & GRU & 99.1 & 91.1 & 98.6 & \textbf{68.1} \\
      & ConvNeXt & \textbf{99.6} & 89.9 & \textbf{98.7} & 64.3 \\
      & Transformer & 99.1 & \textbf{94.5} & 96.1 & 66.9 \\
    \midrule
    \multirow{3}{*}{Gemma 3 4B} & GRU & 96.3 & 69.4 & 94.6 & 49.5 \\
      & ConvNeXt & 98.1 & 75.8 & \textbf{96.1} & 33.7 \\
      & Transformer & \textbf{98.7} & \textbf{90.2} & 96.0 & \textbf{69.1} \\
    \bottomrule
  \end{tabular*}
  \caption{Summation detectors on the teacher's responses and on the backbone's own.}
  \label{tab:teacher}
\end{table}

The measures do not test where a detector places a topic switch, only which topics it uses. Checking boundary placement in decode would need gold labels for generated text, which we leave to future work.

\subsection{RQ3: Do the detectors stay accurate when the response is perturbed?}
\label{sec:rq3}

The same detectors label held-out responses under four perturbations, each scored against its unedited original and passed through the backbone from scratch:
\begin{enumerate}
  \item \textbf{Insertion} of short off-topic asides (jokes or small talk) at topic boundaries, excluded from scoring
  \item \textbf{Removal} of one requested topic from the response, leaving it in the prompt
  \item \textbf{Reordering} of the topic passages so none follows the prompt's order
  \item \textbf{Substitution} of a passage with another on the same topic from a different record, as a control
\end{enumerate}

\begin{figure}[htbp]
  \centering
  \includegraphics{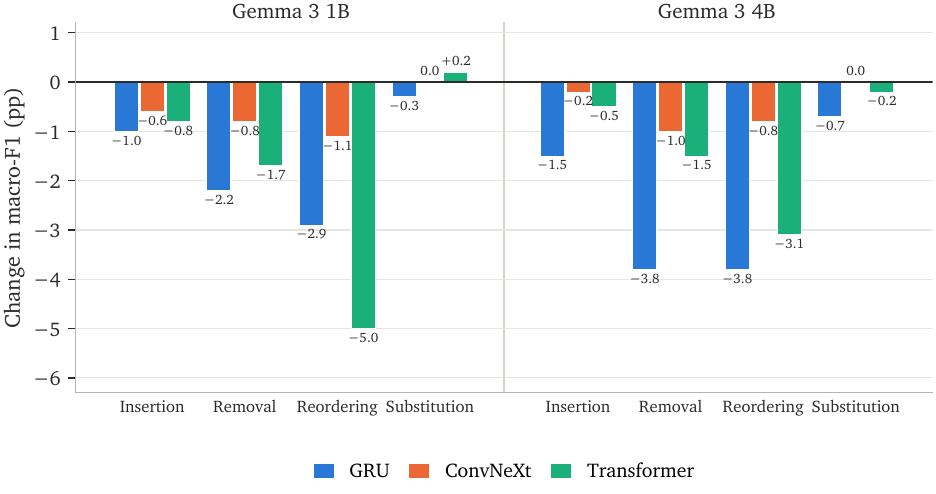}
  \caption{Change in macro-F1 (percentage points) under each perturbation, dynamic features.}
  \label{fig:perturb}
\end{figure}

\textbf{No perturbation reduces macro-F1 by more than 5 points} (\cref{fig:perturb}), and the substitution control leaves scores unchanged, which indicates that the detectors label each passage by its own content. ConvNeXt loses the least under every perturbation, never more than 1.1 points. Reordering affects Transformer the most ($-5.0$ on Gemma 3 1B). It attends to the whole prompt, and every training response answers the topics in the order the prompt asks them, whereas ConvNeXt's 379-token window mostly excludes the prompt.

\begin{figure}[htbp]
  \centering
  \includegraphics{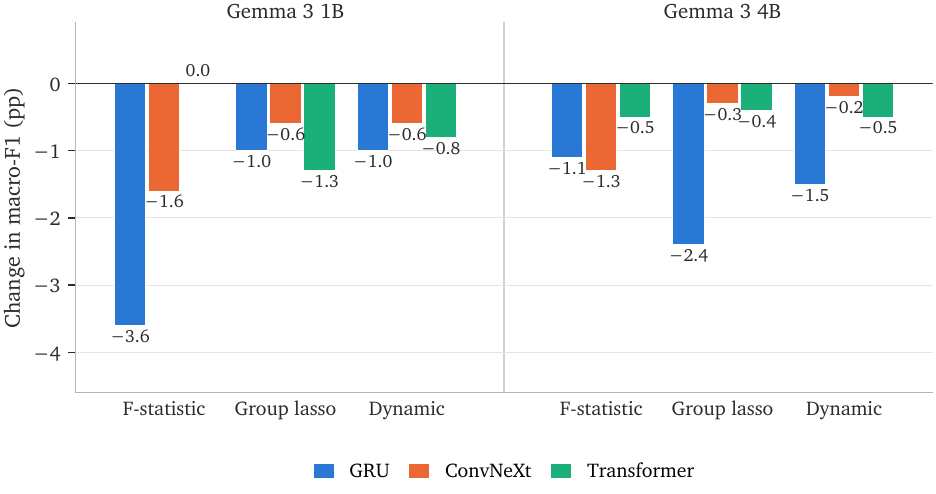}
  \caption{Change in macro-F1 (percentage points) on real tokens after insertion.}
  \label{fig:insertion}
\end{figure}

\paragraph{Insertion across inputs.} The loss on real tokens stays within 3.6 macro-F1 points for each of the three inputs tested, F-statistic, group lasso and dynamic (\cref{fig:insertion}). Most of this loss is precision on General news \& content, the topic closest in content to the inserted asides. For the Gemma 3 1B F-statistic GRU, General news precision falls from 84 to 61 while the other topics move by 1 or 2 points.

\begin{table}[htbp]
  \centering
  \begin{tabular*}{\textwidth}{@{\extracolsep{\fill}}lcccccc@{}}
    \toprule
     & \multicolumn{3}{c}{Gemma 3 1B} & \multicolumn{3}{c}{Gemma 3 4B} \\
    \cmidrule(lr){2-4}\cmidrule(l){5-7}
    Input & GRU & ConvNeXt & Transformer & GRU & ConvNeXt & Transformer \\
    \midrule
    F-statistic & 68.6 & 79.4 & 81.1 & 83.5 & 58.6 & \textbf{89.6} \\
    Group lasso & \textbf{91.0} & \textbf{91.7} & \textbf{86.0} & 80.3 & \textbf{80.6} & 82.4 \\
    Dynamic & 90.7 & 53.7 & 82.2 & \textbf{87.0} & 50.9 & 84.7 \\
    \bottomrule
  \end{tabular*}
  \caption{AUROC (\%) of detector confidence separating real tokens from inserted ones (50 = chance).}
  \label{tab:auroc}
\end{table}

Because the label set has no \enquote{none} class, inserted asides still receive a topic, mostly General news \& content, and a detector can flag them only if its confidence drops. For GRU and Transformer it usually does (AUROC 80--91, \cref{tab:auroc}), but the dynamic ConvNeXt scores 54 and 51, giving inserted asides about the same confidence as real tokens. Its confidence therefore cannot flag inserted text under dynamic features, and group lasso is the only input on which all three detectors, ConvNeXt included, reach 80 at both sizes.

\begin{figure}[htbp]
  \centering
  \includegraphics{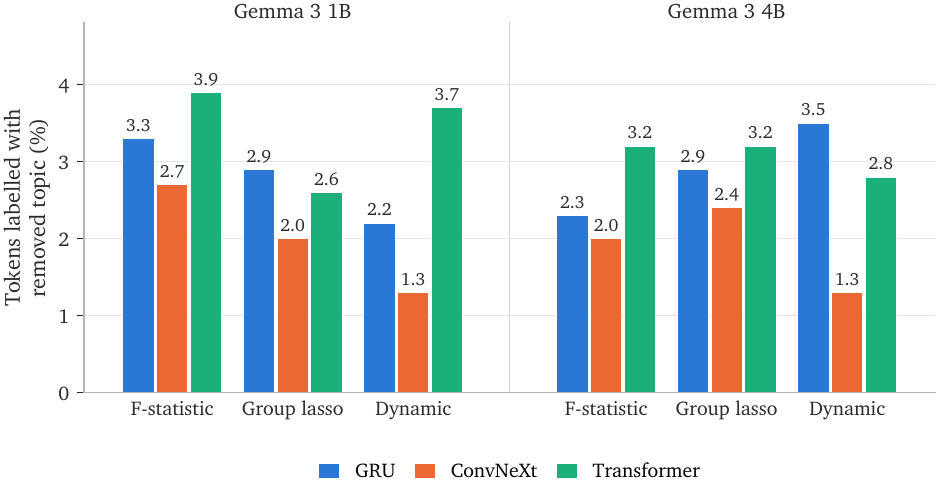}
  \caption{Share of response tokens (\%) labelled with the removed topic.}
  \label{fig:removal}
\end{figure}

\paragraph{Removal.} Averaged over topics, each of these inputs assigns the removed topic to 1 to 4\% of response tokens (\cref{fig:removal}). The rate depends on the topic, from 7 to 8\% for Enterprise documents, the topic hardest to separate in RQ1 and two to three times more affected than any other, down to 1 to 2\% for Healthcare. ConvNeXt assigns the removed topic least often, but does so with 90 to 100\% confidence, consistent with its uniform confidence under insertion.

\paragraph{Reordering.} Under the label-based static rules, Transformer loses 3 to 4 macro-F1 points and ConvNeXt about 1, the same pattern as with dynamic features. Enterprise documents and HR are again the most affected topics, and Healthcare and General news the least.

\subsection{RQ4: Is the SAE necessary?}
\label{sec:rq4}

We repeat RQ1 and RQ2 with the SAEs ablated, training the same detectors on the raw hidden states of the $N$ layers (NoSAE in \cref{tab:rq1,tab:decode}), to see whether the SAE contributes to accuracy, size, or both.

\begin{table}[htbp]
  \centering
  \begin{tabular*}{\textwidth}{@{\extracolsep{\fill}}lrccc@{}}
    \toprule
    Input & Width & GRU & ConvNeXt & Transformer \\
    \midrule
    Static SAE (Summation, F-statistic, Group lasso) & 1,500 & 2.1M & 3.6M & 1.4M \\
    NoSAE, Gemma 3 1B & 11,520 & 9.8M & 6.1M & 4.0M \\
    NoSAE, Gemma 3 4B & 25,600 & 20.7M & 9.7M & 7.6M \\
    Dynamic & $\sim$54,000 & 42.5M & 17.0M & 14.9M \\
    \bottomrule
  \end{tabular*}
  \caption{Detector parameters for each input. Only the input projection changes.}
  \label{tab:params}
\end{table}

\textbf{The SAE is not necessary for accuracy.} Raw hidden states are the best input for Transformer on Gemma 3 1B and ConvNeXt on Gemma 3 4B, and within 0.3 points of the best in three of the remaining four cases. The exception is dynamic ConvNeXt on Gemma 3 1B, which leads NoSAE by 0.6 (97.3 against 96.7). The NoSAE detectors also transfer to decode at least as well, with 95--96 purity for GRU and ConvNeXt and the highest full coverage on Gemma 3 4B.

A static SAE detector, by contrast, has only 1.4M--3.6M parameters at either backbone size (\cref{tab:params}), because its input stays 1,500 wide, while an NoSAE detector's input grows with the model and reaches 7.6M--20.7M on Gemma 3 4B. With label-based selection, this smaller input matches NoSAE for GRU (96.4 against 96.3) and Transformer (95.0 against 94.9) on Gemma 3 4B. ConvNeXt is the exception, where NoSAE leads 97.5 to 95.5. At inference the static stage needs only the selected encoder rows, 1.73M parameters on Gemma 3 1B and 3.84M on Gemma 3 4B, against 378M and 839M for the ten SAEs loaded whole, and every input column is a named feature that can be inspected.

%% file: sections/06_discussion.tex
\section{Discussion and Conclusion}
\label{sec:discussion}

We introduce TopicScope, a framework for training topic monitors on the SAE features of a generating language model, which assign a topic to every token of a response while it is being written. Because the hidden state at a token is computed the same way in prefill and in decode, monitors trained only on teacher-written conversations in prefill transfer without retraining to the responses Gemma 3 1B and 4B generate themselves, labelling 90--98\% of tokens with a requested topic when the features are selected per token or by topic labels. These monitors remain accurate when the response is perturbed, losing at most 5 macro-F1 points when off-topic asides are inserted, a topic is removed or the sections are reordered. Training the same detectors on raw hidden states shows that the SAE is not required for accuracy, but it keeps the monitor small: a static SAE detector has 1.4M--3.6M parameters at either backbone size, whereas a detector on raw hidden states grows with the backbone and reaches 7.6M--20.7M parameters on Gemma 3 4B. Every input column of the SAE detector is also a named feature, which makes the monitor open to inspection in a way raw hidden states are not.

Several directions follow from this work. Decode evaluation would be more complete with gold labels for generated text, which would test where a monitor places each topic switch and not only which topics it assigns. A dedicated off-topic class would let the monitor flag inserted asides directly, without relying on a drop in its confidence. The perturbation analysis could be extended to detectors on raw hidden states, completing the comparison with the SAE inputs. Training on multi-turn conversations collected from deployed assistants would test the monitors on the gradual drift that motivates this work. Larger backbones, other model families and other SAE suites would show whether the fixed monitor size holds as the backbone grows further.

%% file: sections/appendix.tex
\section{Additional prefill results}
\label{app:prefill}

\paragraph{Per-topic overlap.} Enterprise documents and HR \& people operations are the two hardest topics for nearly every detector on Gemma 3 4B (\cref{tab:pertopic}). Both are procedural, internal-process topics with shared vocabulary, since a memo about a staffing policy and an HR plan read alike. A stronger detector also narrows the spread between topics, from 9 overlap points for the summation GRU to 3.5 for the dynamic ConvNeXt.

\begin{table}[htbp]
  \centering
  \begin{tabular*}{\textwidth}{@{\extracolsep{\fill}}lcccccc@{}}
    \toprule
     & \multicolumn{3}{c}{Summation} & \multicolumn{3}{c}{Dynamic} \\
    \cmidrule(lr){2-4}\cmidrule(l){5-7}
    Topic & GRU & ConvNeXt & Transformer & GRU & ConvNeXt & Transformer \\
    \midrule
    Enterprise documents & 72.1 & 83.9 & 80.9 & 79.4 & 93.4 & 84.5 \\
    General news \& content & \textbf{81.3} & \textbf{90.8} & 84.9 & \textbf{90.4} & \textbf{96.4} & 89.5 \\
    Customer service & 79.5 & 86.4 & 86.0 & 86.0 & 93.4 & 89.0 \\
    Legal & 79.3 & 83.1 & \textbf{87.7} & 87.2 & 95.7 & \textbf{91.1} \\
    Financial & 78.1 & 86.9 & 85.4 & 85.4 & 93.9 & 88.6 \\
    HR \& people ops. & 74.1 & 84.7 & 82.2 & 83.5 & 92.9 & 86.8 \\
    Healthcare & 79.6 & 85.8 & \textbf{87.7} & 89.9 & 95.6 & \textbf{91.1} \\
    \bottomrule
  \end{tabular*}
  \caption{Per-topic test overlap (\%) on Gemma 3 4B.}
  \label{tab:pertopic}
\end{table}

\paragraph{Supervising prompt tokens.} Every detector in the main text is trained with labels on both prompt and response tokens. Adding the roughly 515,000 prompt tokens to the 8.2M response tokens leaves response scores nearly unchanged (\cref{tab:supervision}). Ten of the twelve differences are within 1.2 overlap points, and the two larger ones are both static on Gemma 3 4B, where GRU gains 2.4 points and ConvNeXt loses 2.1. GRU likely gains because its running state now receives gradient during the prompt and reaches the response already oriented.

\begin{table}[htbp]
  \centering
  \begin{tabular*}{\textwidth}{@{\extracolsep{\fill}}lccc@{}}
    \toprule
    Input & GRU & ConvNeXt & Transformer \\
    \midrule
    Summation, Gemma 3 1B & 92.2 $\to$ 92.5 & 94.6 $\to$ 94.3 & 92.3 $\to$ 91.1 \\
    Summation, Gemma 3 4B & 79.8 $\to$ 82.2 & 91.0 $\to$ 88.9 & 88.4 $\to$ 87.2 \\
    Dynamic, Gemma 3 1B & 94.5 $\to$ 94.8 & 95.8 $\to$ 96.0 & 90.0 $\to$ 89.8 \\
    Dynamic, Gemma 3 4B & 91.5 $\to$ 90.4 & 96.2 $\to$ 95.7 & 89.6 $\to$ 90.3 \\
    \bottomrule
  \end{tabular*}
  \caption{Overlap (\%) on response tokens, response-only supervision $\to$ prompt-and-response supervision.}
  \label{tab:supervision}
\end{table}

Scores pooled over prompt and response tokens run 1 to 3 points below response-only scores, because short prompt questions give the detector less context than long answers. All main-text scores are computed on the pooled set.

\section{Layer knockout and feature attribution}
\label{app:knockout}

Accuracy alone does not show which layers or features a detector relies on, and a detector that relies on a single layer is exposed to any shift in that layer's signal. We measure this dependence with two methods, applied to every trained detector. In layer knockout, we zero one layer's block of input features, re-score the detector on the test split, and repeat for each of the ten layers. A layer's knockout drop is the overlap lost when it is zeroed. The largest drop gives the detector's worst case, and the sum of the ten drops shows how much it depends on individual layers overall.

In attribution, we compute gradient-times-input \citep{shrikumar2017learning} for the correct-topic logit at each response token, which measures how much each input feature contributed to the correct-topic logit. Within the layer with the largest knockout drop, we then zero the ten highest-ranked and, separately, the ten lowest-ranked features. Finally, we check whether the two methods agree on which layers matter, using the rank correlation between each layer's share of the attribution and its knockout drop.

The worst-layer knockout applies to every input, including NoSAE. The summed knockout and the attribution tests compare how the static rules choose features, so they cover only summation, F-statistic and group lasso.

\begin{figure}[htbp]
  \centering
  \includegraphics{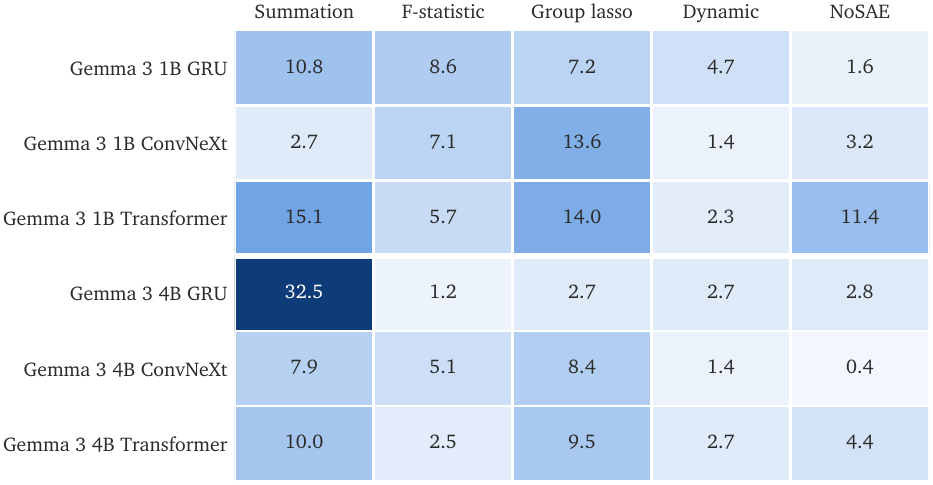}
  \caption{Overlap lost (percentage points) when the single most important layer is zeroed.}
  \label{fig:knockout_worst}
\end{figure}

The selection rule changes how much a detector depends on any one layer. Zeroing layer 29 alone reduces the overlap of the Gemma 3 4B summation GRU by 32.5 points, whereas under the F-statistic its most important layer accounts for only 1.2. Dynamic features and NoSAE spread the dependence most evenly, with no detector losing more than 4.7 points to one layer apart from the Gemma 3 1B NoSAE Transformer (11.4).

\begin{figure}[htbp]
  \centering
  \includegraphics{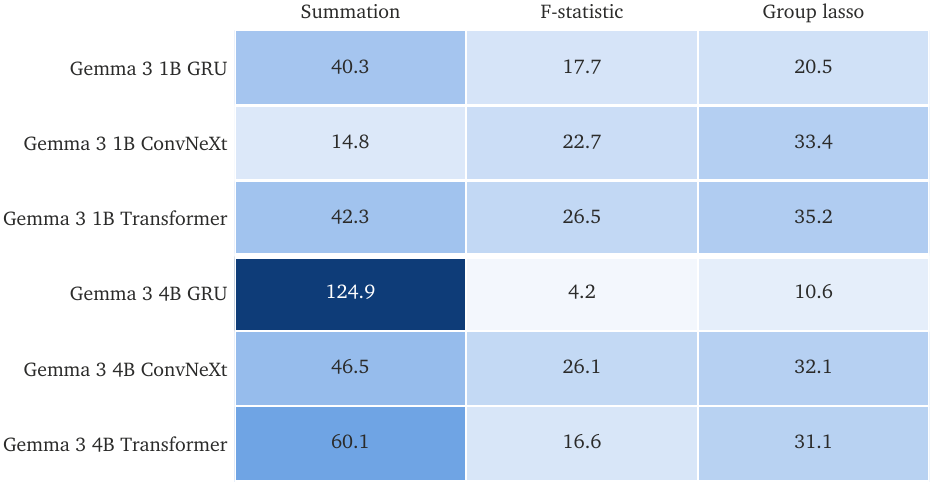}
  \caption{Overlap lost (percentage points), summed over the ten single-layer knockouts.}
  \label{fig:knockout_sum}
\end{figure}

Label-based selection lowers the summed knockout for five of the six detectors, most sharply for the Gemma 3 4B GRU (124.9 to 4.2). The exception is the Gemma 3 1B ConvNeXt, whose total rises to 33.4 under group lasso, almost all of it on layer 16, while its score rises slightly too. The rule also changes which layers a detector draws on. Summation relies mostly on late layers, while group lasso on Gemma 3 1B moves all three detectors onto layer 16, the earliest of the ten.

Within a layer, attribution identifies the features a detector uses. Zeroing the ten highest-attributed features of the most important layer costs 0.1 to 10.3 overlap points, against at most 0.16 for the ten lowest, and the top ten cost more in all 18 detector-and-rule combinations. Under group lasso, zeroing the 75 lowest-attributed features costs at most 0.8 points, so about half of each layer's budget contributes little. For dynamic features the attribution is spread thinly, with the top 100 of about 54,000 columns holding 6 to 16\% of the total.

\begin{figure}[htbp]
  \centering
  \includegraphics{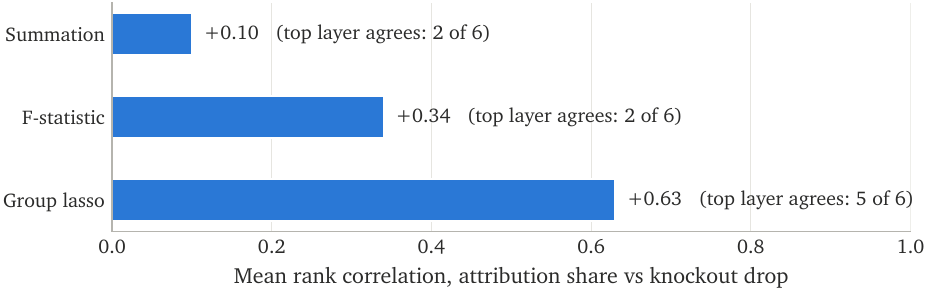}
  \caption{Agreement between attribution share and knockout drop across layers.}
  \label{fig:agreement}
\end{figure}

Across layers, attribution is less dependable (\cref{fig:agreement}). Attribution sums absolute contributions, so a layer with large activations receives a large share whether or not the detector depends on it. Summation picks features by the size of their activations, so it inflates exactly those layers. We therefore rank layers by knockout.

\section{Detector architectures and training details}
\label{app:detectors}

All three detectors read the same per-token input and emit logits over the seven topics at every position. Every detector projects its input to width 256 and ends in a linear head over the seven topics, and all three share the training settings in \cref{tab:archs}. Each is trained with token-level cross-entropy over the labelled tokens, and the checkpoint with the best validation macro-F1 is kept. The body of each detector is the same across inputs, so the parameter counts differ only through the input projection (\cref{tab:params} in the main text). The input width itself is set by each backbone's feature-extraction settings, listed in \cref{tab:extraction}.

\begin{table}[!htbp]
  \centering
  \begin{tabular*}{\textwidth}{@{\extracolsep{\fill}}lccc@{}}
    \toprule
    Hyperparameter & GRU & ConvNeXt & Transformer \\
    \midrule
    Layers / blocks & 3 & 6 & 2 \\
    Hidden width & 256 & 256 & 256 \\
    Feed-forward width & -- & 1,024 & 512 \\
    Attention heads & -- & -- & 8 \\
    Kernel size & -- & 7 & -- \\
    Dilations & -- & 1--32 & -- \\
    Context (tokens) & all & 379 & all \\
    Layer scale & -- & 0.01 & -- \\
    Positional encoding & -- & -- & none \\
    \midrule
    Learning rate & $3 \times 10^{-4}$ & $1 \times 10^{-3}$ & $3 \times 10^{-4}$ \\
    Optimiser & AdamW & AdamW & AdamW \\
    Weight decay & 0.01 & 0.01 & 0.01 \\
    Batch size & 16 & 16 & 16 \\
    Epochs & 15 & 15 & 15 \\
    Dropout & 0.1 & 0.1 & 0.1 \\
    Gradient clipping & 1.0 & 1.0 & 1.0 \\
    Seed & 42 & 42 & 42 \\
    \bottomrule
  \end{tabular*}
  \caption{Detector hyperparameters.}
  \label{tab:archs}
\end{table}

\begin{table}[!htbp]
  \centering
  \begin{tabular*}{\textwidth}{@{\extracolsep{\fill}}lcc@{}}
    \toprule
    Setting & Gemma 3 1B & Gemma 3 4B \\
    \midrule
    Layers read & 16--25 & 24--33 \\
    Hidden width & 1,152 & 2,560 \\
    SAE & Gemma Scope 2 & Gemma Scope 2 \\
    SAE variant & \texttt{l0\_small} & \texttt{l0\_small} \\
    SAE features per layer & 16,384 & 16,384 \\
    Activation & JumpReLU & JumpReLU \\
    \midrule
    $K$ per layer & 150 & 150 \\
    Static width (Summation, F-statistic) & 1,500 & 1,500 \\
    Static width (Group lasso) & 1,438 & 1,489 \\
    Dynamic columns per layer (max) & 6,000 & 6,000 \\
    Dynamic width & $\sim$54,000 & $\sim$54,000 \\
    Dynamic non-zero per token & $\sim$1,380 & $\sim$1,380 \\
    \bottomrule
  \end{tabular*}
  \caption{Feature extraction settings.}
  \label{tab:extraction}
\end{table}